\documentclass[runningheads]{llncs}
\usepackage{amssymb}
\usepackage[colorlinks,linkcolor=red,anchorcolor=black,citecolor=black]{hyperref}
\usepackage[misc]{ifsym}
\usepackage[T1]{fontenc}
\usepackage{graphicx}
\usepackage{color}

\begin{document}
\title{Key-frame Guided Network for Thyroid Nodule Recognition using Ultrasound Videos}

\titlerunning{Key-frame Guided Network}

\author{Yuchen Wang\inst{1} \and
Zhongyu Li\inst{1}$^{(\textrm{\Letter})}$ \and
Xiangxiang Cui\inst{1} \and 
Liangliang Zhang\inst{1} \and 
Xiang Luo\inst{1} \and 
Meng Yang\inst{2} \and 
Shi Chang\inst{3} $^{(\textrm{\Letter})}$} 

%
\authorrunning{Y. Wang et al.}
%
\institute{School of Software Engineering, Xi'an Jiaotong University, China \\
\email{zhongyuli@xjtu.edu.cn} \and
Frontline Intelligent Technology (Nanjing) Co.,Ltd., China \and
Department of General Surgery, Xiangya Hospital, Central South University, China
\email{changshi@csu.edu.cn} }
\maketitle              

\begin{abstract}
Ultrasound examination is widely used in the clinical diagnosis of thyroid nodules (benign/malignant). However, the accuracy relies heavily on radiologist experience. Although deep learning techniques have been investigated for thyroid nodules recognition. Current solutions are mainly based on static ultrasound images, with limited temporal information used and inconsistent with clinical diagnosis. This paper proposes a novel method for the automated recognition of thyroid nodules through an exhaustive exploration of ultrasound videos and key-frames. We first propose a detection-localization framework to automatically identify the clinical key-frame with a typical nodule in each ultrasound video. Based on the localized key-frame, we develop a key-frame guided video classification model for thyroid nodule recognition. Besides, we introduce a motion attention module to help the network focus on significant frames in an ultrasound video, which is consistent with clinical diagnosis. The proposed thyroid nodule recognition framework is validated on clinically collected ultrasound videos, demonstrating superior performance compared with other state-of-the-art methods.

\end{abstract}



\section{Introduction}
The incidence of thyroid cancer has continued to increase during the past decades, which is mainly due to the improvement of detection and diagnosis techniques \cite{yu2020lymph}. Generally, thyroid nodules can be divided into two categories: benign and malignant. Ultrasound examinations are usually used in the diagnosis of thyroid nodules. However, the traditional ultrasound examination process relies heavily on radiologist experience. The diagnostic process is time-consuming and labor-intensive. Besides, overdiagnosis and overtreatment of thyroid nodules have become a global consensus. Therefore, computer-aided diagnosis (CAD) of thyroid nodules from ultrasound examination is of great significance. \par
In recent years, many efforts have been made for the CAD of nodules. Hand-craft features are firstly used in nodule diagnosis, including Chang et al. \cite{chang2010application} extract numerous textural features from nodular lesions and adopt support vector machines (SVM) to select important textural features to classify thyroid nodules. Iakovidis et al. \cite{iakovidis2010fusion} propose a kind of noise-resistant representation for thyroid ultrasound images and use polynomial kernel SVM to classify encoded thyroid ultrasound images. Recently, deep-learning based approaches have been investigated in nodule diagnosis. These works can be summarized into two types by the modality of data. The first type uses static ultrasound images to classify nodule status. For example, Chi et al. \cite{2004Thyroid} use a pre-trained GoogLeNet for feature extraction. The extracted features are then sent to a cost-sensitive random forest classifier to classify nodule status. Li et al. \cite{li2019diagnosis} employ an ensembled ResNet-50 \cite{he2016deep} and Darknet-19 \cite{redmon2017yolo9000} model to recognize malignant thyroid nodules, the model achieved higher performance in identifying thyroid cancer patients versus skilled radiologists. Song et al. \cite{song2018multitask} proposed a two-stage network to detect and recognize thyroid nodules. The second type of method uses ultrasound videos to classify nodule status. For example, Wan et al. \cite{wan2021hierarchical} proposed a hierarchical framework to differentiate pathological types of thyroid nodules from contrast-enhanced ultrasound (CEUS) videos. Chen et al. \cite{chen2021domain} also adopt CEUS videos to classify breast nodules, in which they adopt 3D convolution and attention mechanism to fuse temporal information of the nodule. \par
Despite the above methods have been developed, there are mainly three challenges in the CAD of thyroid nodules. Firstly, most methods achieve the classification of thyroid nodules simply based on static 2D ultrasound images. While in clinical diagnosis, radiologists usually achieve diagnostic conclusions based on ultrasound videos, especially through dynamic morphologies and structures of nodules. Secondly, for most methods considering ultrasound videos, they rely on CEUS videos to get clear foreground and background, which are not the most commonly used B-scan ultrasound. Finally, there are no public datasets available for ultrasound video-based thyroid nodule recognition. Especially, it's hard to collect thyroid ultrasound videos with a unified acquisition process by different radiologists. \par

In this paper, we propose a novel framework for the CAD of thyroid nodules using ultrasound videos. We first develop an automated key-frame localization method to localize the frame with clinically typical thyroid nodules in dynamic ultrasound videos by training a detection-localization model. The model temporally considers texture features of each detected nodule. Subsequently, a key-frame guided network is designed for the classification of ultrasound videos. A motion attention module is introduced into the network to specifically help the network focus on the significant frames in an ultrasound video. Moreover, we collected more than $3000$ clinical thyroid ultrasound videos with a unified acquisition process by three radiologists. The proposed framework is validated in the clinical collected dataset, showing promising performance in both key-frame localization and thyroid nodule classification. \par

\section{Method}
\begin{figure}
\includegraphics[width=\textwidth]{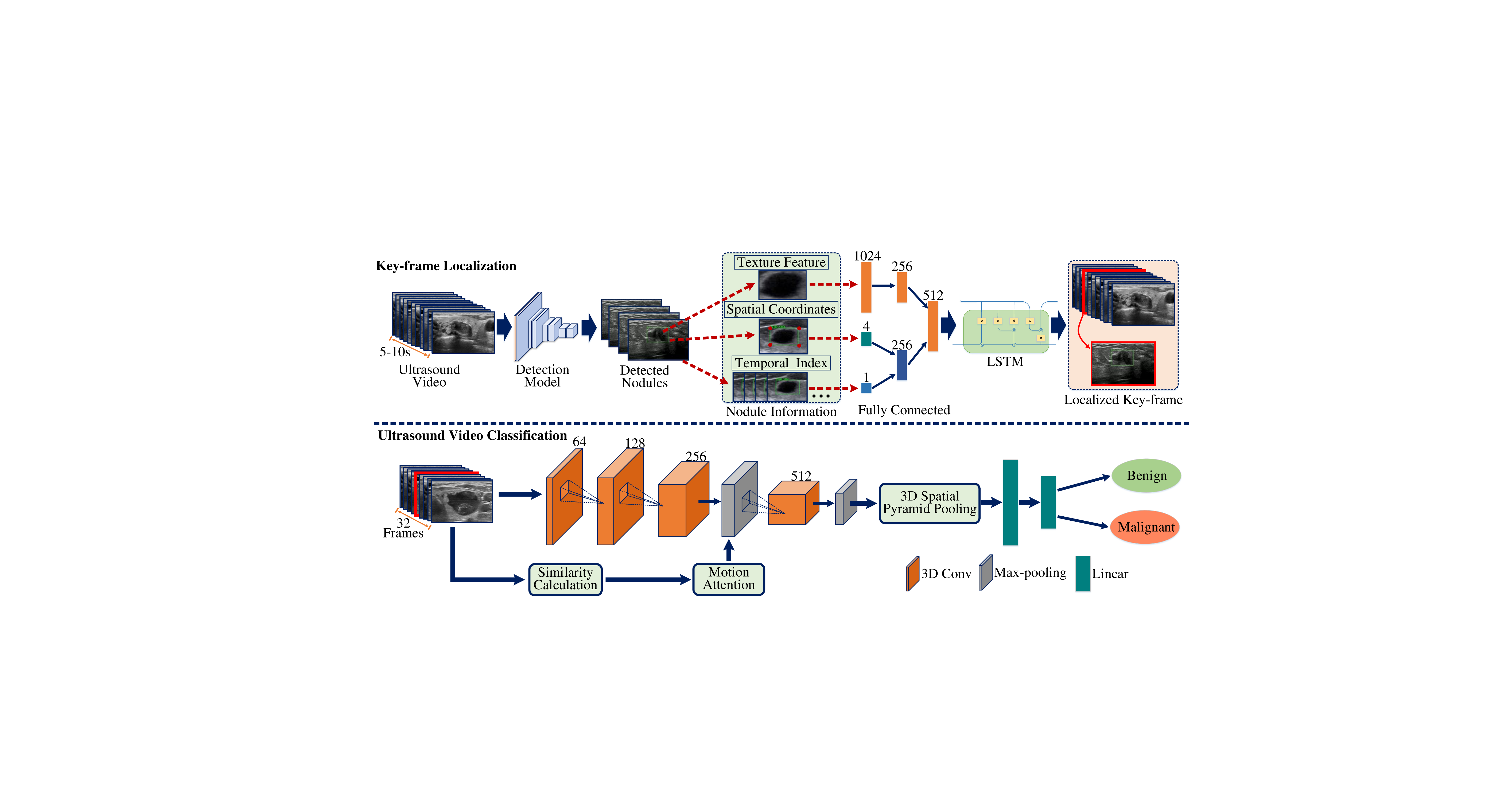}
\caption{Overview of our proposed framework for the thyroid nodule classification in ultrasound videos, which consists of two stages: key-frame localization stage and ultrasound video classification stage.} \label{fig1}
\end{figure}

\textbf{Overview:} As shown in Fig.~\ref{fig1}, we propose a novel framework for thyroid nodule diagnosis using ultrasound videos. Our model consists of two stages: key-frame localization stage and ultrasound video classification stage. In the first stage, we identify the position of the key-frame with clinically typical nodules in thyroid ultrasound videos through a detection-localization pipeline. Given an ultrasound video, thyroid nodules in each frame are first detected. Then the key-frame is automatically localized by simultaneously considering detected nodules with texture, spatial and temporal features. After the key-frame localization, we employ a modified C3D network for the classification of re-organized ultrasound videos with the key-frame centered. Especially, a motion attention module is introduced in considering different importance of each frame for the final classification. The whole framework is automated and can investigate information of spatial, temporal, and key-frame of ultrasound video for the classification of thyroid nodules, which is consistent with clinical diagnosis. \par

\noindent\textbf{Key-frame Localization:} Previous work \cite{jafari2021u} has shown that the key-frame is useful in ultrasound video analysis. We notice that radiologists usually need to localize a key-frame with a typical thyroid nodule in supporting the diagnosis and quantitative analysis. Therefore, key-frame localization in ultrasound videos is important in the CAD of thyroid nodules. To utilize this pattern, we firstly propose a new task to localize the key-frame in a thyroid video, where we design a detection-localization framework to localize the key-frame automatically. Specifically, the network structure includes a detection model which detects nodules in each frame and an LSTM network to predict the index of the key-frame in an ultrasound video, i.e., a sequence of images. \par
For our detection model, we use the Faster-RCNN \cite{ren2015faster} model for nodule detection and feature extraction in each frame. To obtain features of each detected nodule, we take the output of the FC7 layer in Faster-RCNN as the feature representation for each detected nodule. Then, we use a fully connected (FC) layer to reduce the feature dimension to 256. For each detected nodule, we will also need spatial and temporal information for nodule representation in a video sequence. Inspired by Zhou et al. \cite{zhou2019grounded}, we use a 5-D vector to represent spatial and temporal information, including four values for the normalized region of interest (ROI) coordinates and one value for the normalized frame index. Then, the 5-D feature is also projected to a 256-D embedding using an FC layer. Finally, we concatenate the two parts to form a 512-D vector which represents both image features and the spatial and temporal embedding for each nodule. The 512-D vector is then fed into the LSTM network to learn temporal relations. Finally, the output of the LSTM network is sent into two FC layers which eventually output a 1-D score for each nodule. \par
Considering we only have the key-frame index labeled by radiologists, it could be difficult for the network to learn the relationship between ultrasound video and the key-frame index. Therefore, we introduce a regression problem. Specifically, we give each frame a score varying from 0 to 1 to become the key-frame for the network to learn. A label generation procedure is used to make the single key-frame index label into a score sequence label. The score label of the key-frame is set to 1, and the score label of other frames is set from 0 to 1 by calculating the similarity of the current frame and the key-frame. To take full advantage of video data, the similarity includes three parts: image-level similarity, frame index similarity and nodule area IOU (intersection over union) similarity. For image-level similarity, we first calculate the euclidean distance of two frames based on the feature vector extracted in Faster-RCNN. Then we unify them to 0 to 1. For frame index similarity, we calculate the frame index distance between the two frames and unify them to 0 to 1. For IOU similarity, we calculate the IOU of two nodule areas from two frames. Eventually, we average three parts of similarity as our total similarity. Then, the calculated total similarity is used as the score label of each frame for the network to predict. Mean squared error (MSE) loss is adopted for the output score sequence of the network and the score labels. \par

\textbf{Ultrasound Video Classification:} With the key-frame obtained from the previous stage, we design a key-frame guided ultrasound video classification network for thyroid nodule classification. In this stage, we firstly consider using a C3D backbone for nodule classification. However, due to the limited size of our dataset, the C3D network has encountered severe overfitting on our training data. We finally follow the approach in \cite{chen2021domain}. Specifically, we remove most of the parameters in the C3D network to construct our lightweight C3D, which could alleviate overfitting. The architecture of the lightweight C3D network backbone includes four 3D convolutional layers, two max-pooling layers, four batch-normalization layers and two fully connected layers. All the 3D convolutional layers are followed by a batch-normalization layer. In addition, the third and the fourth 3D convolutional layer both have a pooling layer behind them. Among them, the first and the second 3D convolutional layer has a stride of 1$\times$1$\times$1 and 1$\times$2$\times$2, respectively, while the third and the fourth 3D convolutional layers have a stride of 2$\times$2$\times$2. All 3D convolutional layers have a kernel size of 3$\times$3$\times$3 and padding of 1$\times$1$\times$1. The kernel size of max-pooling layers is 2$\times$2$\times$2. \par
By recalling the actual clinical diagnosis of thyroid nodules, radiologists usually determine nodule status by a key-frame to see texture features and its several adjacent frames to see the dynamic morphological change of the thyroid nodule. To incorporate this pattern, we use the key-frame obtained from the previous stage to guide the ultrasound video classification stage. Specifically, to take full advantage of the key-frame, we select $ T $ continuous frames with the key-frame in the center of the sequence as our input rather than uniformly sampled frames in other work \cite{chen2021domain}. $ T $ is empirically set to 32 in this work. Besides, we apply the 3D version of the spatial pyramid pooling (SPP) layer \cite{he2015spatial} to learn multiscale representation. The 3D SPP layer is behind the last pooling layer. Limited by the size of the feature map, we adopt a three-level pyramid (original feature map, 2$\times$2$\times$2, 1$\times$1$\times$1). Then we concatenate the features to form a multiscale representation of the nodule. Eventually, two FC layers with ReLU activation are appended to give category scores. A dropout layer with the dropout rate of 0.5 is added to alleviate overfitting. We use cross-entropy loss for nodule classification, where the classification loss $ L_{cls} $ is formulated by:
\noindent
\begin{equation}
L_{cls} = -(y_{n}\cdot\log z_{n} + (1-y_{n})\cdot\log (1-z_{n})) 
\end{equation}
where n is the batch size, $ y_{n} $ is the label of this batch and $ z_{n} $ is the output of this batch. \par

\begin{figure}
\begin{center}
\includegraphics[width=0.8\textwidth]{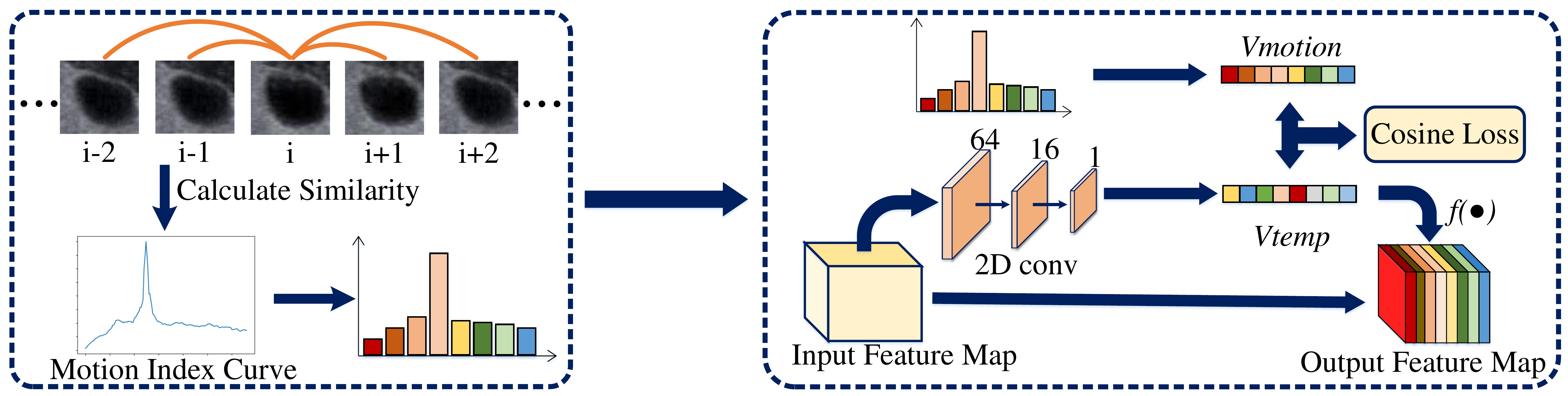}
\end{center}
\caption{The structure of the motion attention branch. It uses image similarity to quantify the motion of radiologists. Then it uses convolutional layers to calculate the temporal weights. Cosine loss is appended between temporal weights $ V_{temp} $ and the motion vector $ V_{motion} $ to optimize parameters.} \label{fig2}
\end{figure}

\noindent\textbf{Motion Attention:} We notice that when radiologists find a possible key-frame, they will slow down their motion to observe the nodule, which means the frames in ultrasound video become relatively still. Therefore we design a motion attention module to perform the attention mechanism similar to the radiologists. We use a structure similar to \cite{chen2021domain} to reassign temporal weights. The motion attention module is added to the end of the first 3D pooling layer. The detailed structure of our motion attention module is shown in Fig.~\ref{fig2}. The structure consists of three 2D convolutional layers. The first and the second 2D convolutional layer have a kernel size of 3$\times$3 with a stride of 2$\times$2, while the last 2D convolutional layer has a kernel size of 2$\times$2 with a stride of 1$\times$1. After these convolutional layers, the feature map is reduced to 1 channel. The final output feature map size is 1$\times$T$\times$1$\times$1, which could be used as temporal weights. On the other side, we split the whole frame sequence into $ T $ (the input temporal dimension) time windows. We use image similarity of these frames to quantify the motion of the radiologist, i.e., high similarity reflects that the radiologist is moving very slowly. We calculate SSIM \cite{wang2004image} and histogram similarity within a small interval (frame $ i-2, i-1, i+1, i+2 $) for frame $ i $. Then we average the results of these two similarity indices to evaluate both pixel-level similarity and structural similarity. We define a motion index to quantify the motion of the radiologist. Motion index $ M_{i} $ for frame $ i $ is defined by the average similarity of frame $ i-2, i-1, i+1, i+2 $. Then we average $ M_{i} $ within each window to represent the motion in this time window. The result vector is defined as $ V_{motion} $. Finally, the learned temporal weights $ V_{temp} $ are multiplied with the input feature map in the temporal dimension. Moreover, consistency loss is applied to optimize the temporal weights. Specifically, our loss function $L_{motion}$ is calculated by the cosine similarity between temporal weights $ V_{temp} $ and the motion vector $ V_{motion} $. The loss function is expressed as: 
\noindent
\begin{equation}
L_{motion} = 1 - cosine(v_{temp}, v_{motion})
\end{equation}
where $ v_{temp} $ is the temporal weights, $ v_{motion} $ is the calculated motion index vector. Finally, we add $ L_{motion} $ to the classification loss $ L_{cls} $, the overall loss $ L $ for our nodule classification network could be written as:
\noindent
\begin{equation}
L = L_{cls} + L_{motion} 
\end{equation}

\section{Experiments and Results}
\noindent\textbf{Dataset:} Our ultrasound video dataset was collected from 2020/04 to 2021/12 at three medical centers, following data cleaning to filter out videos with bad quality. We cropped videos to remove device and patients’ information. The devices include three types, i.e., SAMSUNG MEDISON H60, HS50, and X60. All three devices have line array probes with a frequency of 7.5MHz. In the experiment, we only use videos captured in the cross-section direction of the thyroid on thyroid left or right sides. All videos were annotated with key-frame index, nodule ROI position and nodule status (benign/malignant) by two radiologists with over ten years’ experience. All the annotations were checked by a third radiologist with over twenty years’ experience. Difficult/disagreement samples were decided by three experts together. Finally, 3668 thyroid ultrasound videos were used as our dataset for the key-frame localization stage. For the training of our detection model, we extracted 23219 thyroid images from these videos to train the Faster-RCNN. All images are resized to 500$\times$600 for detection. Then we use these 3668 videos to train the key-frame localization model. We randomly choose 2648 videos for training, 504 videos for validation and 516 videos for testing. The code and data are available at \url{https://github.com/NeuronXJTU/KFGNet}. \par
In the ultrasound video classification stage, due to the extremely unbalanced numbers of benign and malignant nodules, we choose 244 videos from the test set of the previous key-frame localization stage as our dataset, including 125 benign nodule videos and 119 malignant nodule videos. The input of our ultrasound video classification stage is 112$\times$112$\times$32. We resize each nodule to 112$\times$112 and choose a 32-frame sequence for inputs as mentioned before. In addition, data augmentation including random flip and intensity shift is applied to mitigate overfitting. Considering the limited size of our dataset, 5-fold cross validation is applied to verify our model. \par
\textbf{Settings:} The model is implemented with PyTorch 1.5.0, using an RTX 2080 Ti GPU. In our experiment, the whole model is randomly initialized with no pretraining. The two stages are trained separately. In the key-frame localization stage, we first train our Faster-RCNN using default settings in the framework \cite{jjfaster2rcnn}. Then we train the LSTM network using Adam optimizer with batch size 64, where the learning rate is set as 0.01 for 20 epochs training. We use the accuracy of different frame distance tolerance to evaluate our model. Specifically, for frame distance $ D $, if the distance between the predicted key-frame and the key-frame label is smaller than $ D $, we define this as a positive sample. Otherwise, it is defined as a negative sample. Then we calculate accuracy to show the prediction results. For the training of ultrasound video classification stage, we use Adam optimizer with the learning rate of 1e-3 and weight decay of 1e-8. Batch size is set as 16. After training for 20 epochs, we change the learning rate to 1e-4 to train another 20 epochs. Then, accuracy, sensitivity, specificity, precision and F1-score are used to evaluate our ultrasound video classification stage. \par

\textbf{Experimental Results:} To evaluate the proposed method, we conduct a series of comparison and ablation experiments. Since our model contains two stages, we will first evaluate the key-frame localization stage, and then we will evaluate the whole model. \par

\begin{figure}
\includegraphics[width=\textwidth]{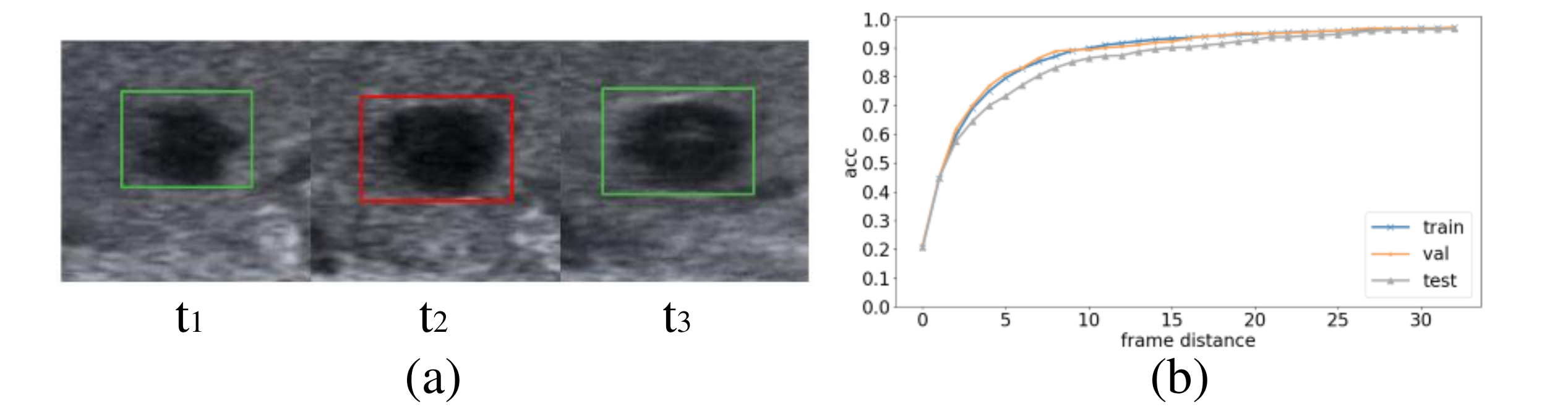}
\caption{Results of key-frame localization stage. (a) Detection results of a few frames in a thyroid ultrasound video, where $ t_{2} $ is the key-frame. (b) Accuracy of key-frame localization with different frame distance tolerance (from 0 to 32).} \label{fig3}
\end{figure}

\noindent For our key-frame localization stage, our Faster-RCNN model reaches AP50 of 77.34\%, which could verify the effectiveness of our detection model. The key-frame localization results are shown in Fig.~\ref{fig3}(b). We calculate the distance between our predicted key-frame and the key-frame label. The results show accuracy with different frame distance tolerance (from 0 to 32). According to the results, our model can reach more than 70\% of accuracy within 5 frames error and can reach more than 90\% of accuracy within 15 frames error, which is significant for the next ultrasound video classification stage. \par
We then evaluate the whole model on our dataset. Table~\ref{tab1} shows the results of different methods and our ablation experiments. Since we are using thyroid ultrasound video data, we choose some representative video classification models for comparison, such as C3D \cite{tran2015learning}, R3D \cite{tran2018closer} and R2plus1D \cite{tran2018closer}. We choose lightweight C3D as our baseline because our further improvements are based on the lightweight C3D network. The results show that our lightweight C3D backbone achieves the highest accuracy of 68.45\% compared with other video classification backbones and gets competitive results in other metrics. Moreover, we design our ablation experiments to verify the effectiveness of the key-frame, 3D SPP and motion attention module, respectively. We first validate the effectiveness of the key-frame. The results show that key-frame guided lightweight C3D outperforms our baseline in all metrics. Specifically, our model has improved 3.2\% in accuracy by key-frame, which is mainly because our key-frame has alleviated temporal information redundancy and helped the network focus on significant frames. We then validate 3D SPP and motion attention, respectively. Notably, since the key-frame is our fundamental design and our motion attention module is closely related to the key-frame, we will further validate these two modules based on the key-frame by adding them to our key-frame guided lightweight C3D. Results show that both 3D SPP and motion attention could further improve our network performance. Specifically, the accuracy can get a boost of 2.5\% by motion attention and 2.9\% by 3D SPP. Finally, we implement our model with key-frame, 3D SPP and motion attention. According to the results, our proposed method outperforms other existing methods by achieving the highest performance in all metrics. Our proposed network outperforms the baseline by 8.2\% in accuracy and 7.5\% in sensitivity, respectively. \par

\begin{table}
\caption{Quantitative results of our proposed method with the comparison to other state-of-the-art methods and our ablation experiments.}\label{tab1}
\resizebox{\textwidth}{!}{
\begin{tabular}{lcccccccc}
\hline
Methods & \quad Key-frame \quad & \quad 3D SPP \quad & \quad Motion attention \quad & Accuracy & Sensitivity & Specificity & Precision & F1-score \\
\hline
C3D \cite{tran2015learning}        & & &     & 66.78\% & 70.51\% & 65.83\% & 63.27\% & 65.47\% \\
R3D \cite{tran2018closer}        & & &     & 67.19\% & 68.08\% & 68.45\% & 70.55\% & 68.27\% \\
R2plus1D \cite{tran2018closer}   & & &     & 68.45\% & 71.37\% & 66.96\% & 67.83\% & 68.75\% \\
\hline
lightweight C3D        & & &     & 68.45\% & 71.06\% & 66.73\% & 66.52\% & 68.13\% \\
lightweight C3D  &\checkmark & &     & 71.69\% & 71.94\% & 71.88\% & 74.73\% & 72.76\% \\
lightweight C3D  &\checkmark & &\checkmark & 74.19\% & 77.88\% & 72.04\% & 71.59\% & 73.90\% \\
lightweight C3D  &\checkmark &\checkmark & & 74.59\% & 76.67\% & 75.25\% & 76.69\% & 75.52\% \\
Ours  &\checkmark &\checkmark &\checkmark & \textbf{76.65\%} & \textbf{78.52\%} & \textbf{76.36\%} & \textbf{77.17\%} & \textbf{77.08\%} \\
\hline
\end{tabular}}
\end{table}

\section{Conclusions}
In this paper, we propose a novel method for computer-aided diagnosis of thyroid nodules using ultrasound videos. To the best of our knowledge, this is the first time to achieve the automated localization of key-frames with typical thyroid nodules in ultrasound videos. The proposed method can locate the key-frame in a thyroid video and use the key-frame to guide the nodule classification. Besides, the motion attention mechanism is introduced to help the network focus on significant frames in a video. As a result, all the designs improve the performance of our network. Our results indicate that the proposed method has significantly outperformed other state-of-the-art methods for thyroid nodule classification on our dataset. The proposed framework can be applied to clinical CAD in reducing doctors' work and improving diagnostic accuracy. \par

\subsubsection{Acknowledgements} This work is partially supported by the National Natural Science Foundation of China under grant No. 61902310 and the Natural Science Basic Research Program of Shaanxi, China under grant 2020JQ030.

\bibliographystyle{splncs04}
\bibliography{mybibliography}

\end{document}